# Rapidly Developing High-quality Instruction Data and Evaluation Benchmark for Large Language Models with Minimal Human Effort: A Case Study on Japanese

Yikun Sun[♦], Zhen Wan[♦], Nobuhiro Ueda
Sakiko Yahata, Fei Cheng, Chenhui Chu, Sadao Kurohashi

Kyoto University
{sun, zhenwan, ueda, yahata}@nlp.ist.i.kyoto-u.ac.jp
{feicheng, chu, kuro}@i.kyoto-u.ac.jp

## Abstract

The creation of instruction data and evaluation benchmarks for serving Large language models often involves enormous human annotation. This issue becomes particularly pronounced when rapidly developing such resources for a non-English language like Japanese. Instead of following the popular practice of directly translating existing English resources into Japanese (e.g., Japanese-Alpaca), we propose an efficient self-instruct method based on GPT-4. We first translate a small amount of English instructions into Japanese and post-edit them to obtain native-level quality. GPT-4 then utilizes them as demonstrations to automatically generate Japanese instruction data. We also construct an evaluation benchmark containing 80 questions across 8 categories, using GPT-4 to automatically assess the response quality of LLMs without human references. The empirical results suggest that the models fine-tuned on our GPT-4 self-instruct data significantly outperformed the Japanese-Alpaca across all three base pre-trained models. Our GPT-4 self-instruct data allowed the LLaMA 13B model to defeat GPT-3.5 (Davinci-003) with a 54.37% win-rate. The human evaluation exhibits the consistency between GPT-4's assessments and human preference. Our high-quality instruction data and evaluation benchmark are released here[1,2].

**Keywords:** instruction tuning, evaluation benchmark, self-instruct, GPT-4 assessment

## 1. Introduction

Recent advancements in large language models (LLMs) have aimed to refine their capacity to accurately follow human instructions and navigate intricate scenarios (Chiang et al., 2023; Taori et al., 2023; Peng et al., 2023; Wang et al., 2023a). This target can be accomplished by supervised fine-tuning (SFT) using manually (Databricks, 2023; Köpf et al., 2023) or automatically (Wang et al., 2023b) generated instruction data. A prevalent method in this domain is self-instruction (Wang et al., 2023b; Honovich et al., 2022; Xu et al., 2023), which generates English SFT data with a state-of-the-art (SOTA) LLM, such as GPT-4 (OpenAI, 2023) to train open-source LLMs. This method proves to be an effective and efficient way to strengthen open-source LLMs with minimal human effort. However, a notable limitation of current methodologies is their predominant focus on English, leading to the gap in the abundance of data resources between English and other languages, such as Japanese.

Recent work (rinna, 2023; Kunishou, 2023) attempts to solve this gap by using machine translation to convert an English instruction dataset to Japanese. For instance, Japanese-Alpaca (Kunishou, 2023) is fine-tuned with the instruction data translated from English Alpaca via GPT-3.5 (Brown et al., 2020). We suspect that the original English data is generated by outdated LLMs such as GPT-3.5, and translation could diminish the quality of instruction.

In this paper, we propose a novel method for generating Japanese instruction with GPT-4 directly. We translated the original English manual seed instruction tasks (Taori et al., 2023) into Japanese. Native Japanese speakers proofread the translations to ensure native-level fluency and natural expression. The self-instruction method was then employed, utilizing GPT-4 to generate diverse instruction data. This self-instruct method generated data notably outperformed the data that was directly translated from English in Table 4.

Besides the lack of instruction data, another gap exists in the shortage of resources for evaluating Japanese LLMs. How to evaluate LLMs re-

---

[♦] denotes equal contribution.
[1] https://github.com/hitoshizuku7/awesome-Ja-self-instruct
[2] https://github.com/ku-nlp/ja-vicuna-qa-benchmark

mains an open question. An effective and popular approach is to construct question-answering pairs of various aspects (e.g., knowledge, writing, coding, etc.) to assess the capabilities of LLMs comprehensively. However, constructing such evaluation benchmarks involves human efforts to create reference answers, which is costly and time-consuming.

Encouragingly, an increasing number of studies (Chiang et al., 2023; Wang et al., 2023a; Chen et al., 2023) have revealed that LLMs like GPT-4 show the reliable capability to assess model outputs in a reference-free manner, i.e., no need for human reference. To solve the second gap, we propose to follow the reference-free trend to construct the evaluation benchmark with $8$ categories of $80$ Japanese questions translated from English Vicuna (Chiang et al., 2023), design Japanese prompts to enhance the evaluation reliability, and let GPT4 be the judge with no reference offered.

In summary, existing two gaps largely influence the fast development of LLMs in non-English languages. In this paper, we focus on a case study on Japanese to build a bridge between the two above gaps between English and Japanese, with minimal human effort. We conclude our contributions as follows:

- High-quality Japanese instruction dataset generated by GPT-4 self-instruct for LLM SFT.

- Japanese LLMs evaluation benchmark with $8$ categories of $80$ Japanese questions translated by native speakers and assessed by GPT-4 for evaluation .

- Experiment results demonstrate that models fine-tuned on our GPT-4 self-instruct data consistently outperform existing approaches, and human evaluations validate the consistency between GPT-4's assessments and human preferences.

## 2. Related Work

### 2.1. Instruction Data Generation

The success of instruction tuning (Longpre et al., 2023; Wei et al., 2022a; Zhang et al., 2023) requires that the instruction data is sufficiently diverse and representative to unlock LLMs' potential for solving downstream tasks. High-quality instruction datasets are often accumulated by human annotation. Databricks (2023), Köpf et al. (2023) are gathered via crowdsourcing. Recently, Wang et al. (2023b); Taori et al. (2023); Chiang et al. (2023) Wang et al. (2023a) has sparked a trend in automatically generating instruction data by distilling it from other LLMs with proper prompt guidance.

For a non-English language, such as Japanese, due to the high cost and time consumption of manual annotation, a typical approach (rinna, 2023; Kunishou, 2023) is to translate English instruction data into Japanese for fine-tuning Japanese LLMs such as Japanese-Alpaca. We argue the outdated version of the English source and the limits of translation quality of this approach. Therefore, we propose a novel method for generating Japanese instruction with GPT4 directly.

### 2.2. LLM Evaluation

Recently, constructing question-answering pairs of various aspects (e.g., knowledge, writing, coding, etc.) to assess LLMs' capabilities comprehensively has become popular. Several QA benchmarks with manual reference answers have been established (Bai et al., 2022; Geng et al., 2023; Wang et al., 2023b). Leveraging LLMs like GPT-4 to assess model answers in a reference-free manner naturally becomes a trend due to its convenience and efficiency (Chiang et al., 2023; Wang et al., 2023a; Chen et al., 2023). Though studies like Wang et al. (2023a); Min et al. (2022); Zheng et al. (2023) suggest that the judgment of LLMs may exhibit certain biases to answer length and order, these issues can be gradually alleviated with the advancement of prompt research and model performance. We therefore employ a reference-free evaluation scheme utilizing the GPT-4 judge. Our comprehensive manual evaluation reveals that GPT-4's assessments are highly consistent with human preference.

## 3. Methodology

Our methodology consists of two parts. The first part is the generation of Japanese instruction data for LLMs SFT. With them, we can fine-tune various pre-trained models on natural and high-quality instruction data. The second part entails the construction of a credible evaluation benchmark for Japanese LLMs. The following subsections elaborate on the key components of our methodology.

### 3.1. GPT-4 Self-instruct Generation

Figure 1 shows the flow chart of our self-instruct generation method. We revisited the original self-instruct method, which leverages a limited amount of English seed tasks to generalize and generate new task data. Our method translates that small number of seed tasks into Japanese and manually post-edits them to achieve native-level quality. We can then utilize GPT-4 to generate high-quality Japanese data directly. We assume that such native seeds-guided self-instruct data will substantially surpass the quality of translated data.

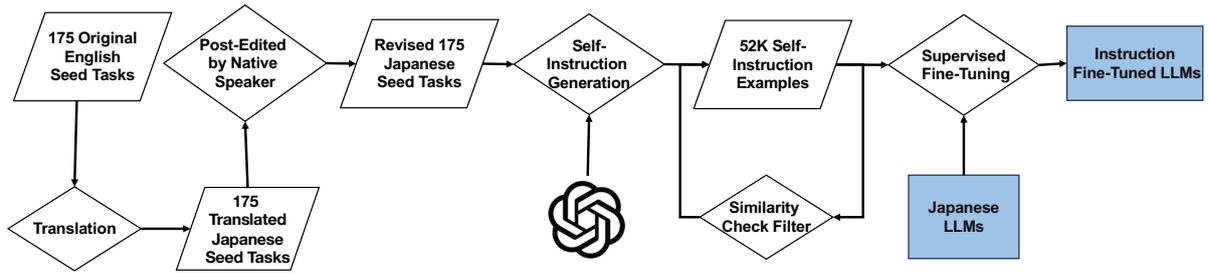

Figure 1: The flow chart of our self-instruct proposal

### 3.1.1. Translate Seed Tasks

To generate a large number of high-quality instruction data through self-instruct, we must ensure that the quality of the seed tasks is sufficiently high. We use the same $175$ human-written instruction seeds as Taori et al. (2023); Wang et al. (2023b). First, we utilized GPT-4 to translate them into Japanese. Then, two native Japanese speakers help to review and post-edit the translations to obtain native-quality Japanese seed tasks.

### 3.1.2. Generate Self-instruct Data by GPT-4

Then, we use the LLM GPT-4 accessed through the OpenAI API to generate the instruction data.

**Construct Japanese self-instruct prompt**
We follow Taori et al. (2023); Wang et al. (2023b) to construct a comprehensive prompt with the following requirements to guide GPT4 to automatically generate new instruction data.

- Instruction Diversity: This requirement encourages the generated instruction with diverse content.

- Instruction Feasibility: It emphasizes restricting the generated instruction, which should be able to be completed by LLMs, rather than necessitating visual or audio instruction.

- Instruction Format: This part defines the format of the generated instruction. For instance, the generating instructions should be in Japanese and comply with certain length constraints.

- Input Section: This requirement clarifies the role of the input section, which serves as a comprehensive supplement to the instruction. Importantly, sometimes it is also acceptable for an instruction data example to lack input section.

- Output Section: It encapsulates the rules of the output section. The output must respond appropriately to the instruction with length limitation.

**Generate instruction data** We utilize GPT-4 to generate the new instruction data in a few-shot manner. At each round of generation, we randomly sample $3$ task examples from seed tasks to append them to the prompt. These task examples will be utilized by GPT-4 as a demonstration to generate instruction data of the same format. The prompt instructs GPT-4 to generate new examples until the completion of a total number of $20$, including $3$ seed tasks and $17$ newly generated examples.

**Check the similarity of the generated data and filter** In order to encourage diversity of the instruction data to ensure the instruction data quality, we need to check the similarity and filter the overly similar instruction examples generated in each round. After each generation round, we use Juman++ (Tolmachev et al., 2018) to do segmentation and use ROUGE-L (Lin, 2004) to assess the similarity of a newly generated instruction against all previously generated data within the instruction data pool. If any data in the instruction data pool with the ROUGE-L score exceeding $0.7$, it signifies that the newly generated instruction lacks sufficient diversity and should be excluded.

In addition to this similarity assessment, the work also employs a blacklist to filter out instruction data that is either unsuitable for SFT or cannot be processed by LLMs. All instruction containing keywords such as audio, video, images, and so forth will be filtered out by the blacklist.

We finally generated the same amount of instruction data as the English instruction data, a total of $52K$ examples.

### 3.2. Evaluation Benchmark

In this section, we describe the process of creating a QA benchmark consisting of $80$ high-quality questions for evaluating Japanese LLMs. The benchmark follows the reference-free evaluation manner, which leverages GPT-4 to assess the quality of the LLM answers.

#### 3.2.1. Obtain Evaluation Question Data

The original questions data set, drawn from the English Vicuna benchmark, is divided into $8$ common question categories of user prompts to guide LLMs response generation. These questions are designed to test instruction-following ability, covering common use cases and focusing on challenging questions to differentiate models. We manually translate these questions into Japanese and proofread them for their quality. We list a common-sense question below.

> 'ホラー映画を見たりジェットコースターに乗ったりと、怖いという感覚を楽しむ人がいる一方で、こうした体験を避ける人がいるのはなぜですか？' (Why do some people enjoy the sensation of being scared, such as by watching horror movies or going on roller coasters, while others avoid these experiences?)

#### 3.2.2. Assess LLM Responses by GPT-4

In this work, each LLM being evaluated needs to answer all of those $80$ Japanese questions. We will use two methods to evaluate these responses.

- Pairwise mode: GPT-4 performs pairwise comparisons of responses from different LLMs to ascertain which one performs better or yields comparable results for a given question (see Figure 2). The LLMs' capabilities can be evaluated by statistically comparing the pairwise models' win/loss/tie rates across $80$ questions.

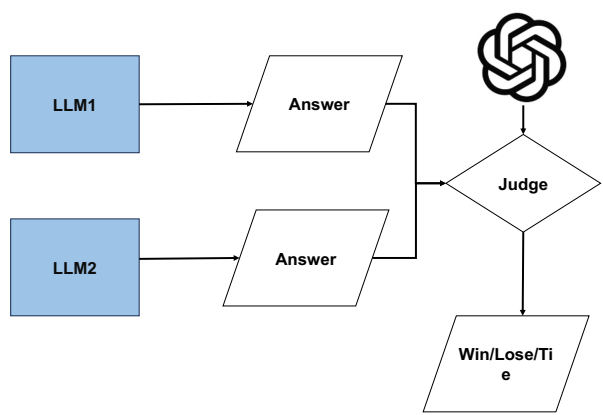

Figure 2: Pairwise mode flow chart

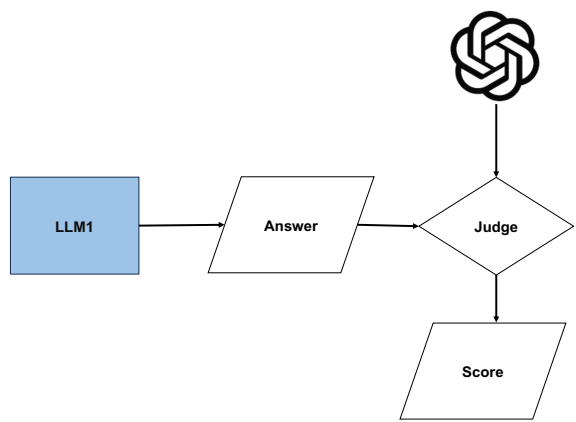

Figure 3: Single score mode flow chart

- Single score mode: GPT-4 directly assigns a score to an answer generated by LLMs (see Figure 3). GPT-4 will judge these answers by considering factors such as the helpfulness, relevance, accuracy, depth, creativity, and level of detail of the response. Then GPT-4 assigns a score ranging from $1$ to $10$. Based on this score, we can directly assume the ability between the different models on specific questions. It is also a method to evaluate the capability of LLMs quantitatively.

## 4. Experiment Setup

We conduct experiments for investigating two research questions: (1) Is the GPT-4 self-instruct data significantly better than the data trans-lated from English Alpaca? (2) How is the performance of our instruction fine-tuned LLMs compared to GPT3.5?

## 4.1. Setting of Fine-tuning Instruction Data

### 4.1.1. Two Sets of Instruction Data

This section aims to demonstrate the superior quality of the instruction data generated by the self-instruct method via GPT4. Two distinct sets of instruction data are deployed for the SFT of various pre-trained models:

- MT Alpaca (baseline): Instruction data is machine-translated from Stanford Alpaca in English. The original English instruction data is self-instructed by GPT-3.5 (Davinci-003), utilizing a seed set of manual instructions.

- Self-instruct (proposed): Utilizing GPT-4-0613 as an instruction-following model, the instruction data were generated by our self-instruct method with high-quality Japanese seed tasks.

We perform SFT with these two sets of instruction data for multiple pre-trained models. Afterward, we can compare the performance of the pair of trained LLMs by two evaluation methods to judge the quality of the instruction data used for SFT.

### 4.1.2. Multiple Pre-trained Models

In this experiment, we choose three pre-trained models for supervised fine-tuning. The target pre-trained models include:

- LLaMA2 7B,13B
- LLaMA 7b
- OpenCALM 7B

LLaMA ([Touvron et al., 2023a](#)) and LLaMA2 ([Touvron et al., 2023b](#)) represent the largest, highest-quality cross-lingual pre-trained models available to the community. OpenCALM ([Andonian et al., 2021](#); [CyberAgent, 2023](#)) comprises a language model pre-trained on Japanese datasets, including Wikipedia and Common Crawl in Japanese. Given this model's pre-training foundation in Japanese data, its performance is anticipated to be superior when SFT with Japanese instruction data.

We leverage the low-rank adaptation (Lora) ([Hu et al., 2021](#)) in our fine-tuning, which proves to be computationally efficient while ensuring the model's performance. For training details, we follow standard Lora hyper-parameters for all models: we fine-tune for $4$ epochs with a learning rate of $5e-5$ and $100$ warmup steps. We also set the Lora rank to be $8$. All models are trained on a single A100 GPU. For the inference, we set a temperature of $0.95$ and max new tokens of $512$ for all models.

In this work, due to laboratory budget constraints, we selected the 7B model as the primary model for SFT and comparison target. Additionally, within laboratory budget constraints, we selected the larger size model, LLaMA2 13B as the pre-trained model, intending to pursue optimal performance for challenging GPT-3.5.

## 4.2. Evaluate the Fine-tuned LLMs with Our Benchmark

After generating answers to the questions in benchmark with language models, we can compare the language capability among the models. In this paper, we implement two lines of evaluation:

- **Ploting learning curve with increasing data:** We designate the instruction data from $1K$, $2K$, $5K$, and every subsequent $10K$ as checkpoint. At that time, we will randomly draw equivalent samples of instruction data from the machine-translated instruction data for SFT. Each checkpoint will then be assessed using the single score mode, and a corresponding learning curve will be plotted for each pre-trained model. The main target of this assessment is to monitor any changes in score throughout the SFT process with increasing size of instruction data.

- **Comparing fine-tuned models to GPT-3.5:** For the comparison of each model after the fine-tuning on $52K$ instruction data, we leverage two modes: the single-score mode to directly score each model's performance and the pairwise mode to compare each model with the GPT-3.5 (davinci-003). We use two modes to enhance the accuracy of our evaluation and at the same time, to analyze the consistency between two modes in evaluation.

## 5. Result Analysis

### 5.1. Plot Learning Curve with the Single-score Mode

As in Figures [4](#), [5](#), and [6](#), we evaluated each checkpoint on $80$ questions in the single mode to observe the performance improvements caused by the increasing instruction data. We also tested the answer scores of the pre-trained model without SFT and concluded that SFT with instruction data will significantly improve the models' language ability and performance.

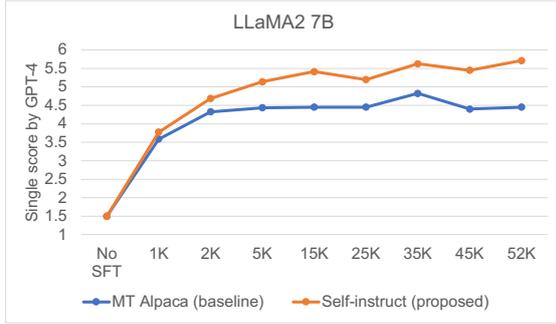

Figure 4: The LLaMA-2 7B learning Curve

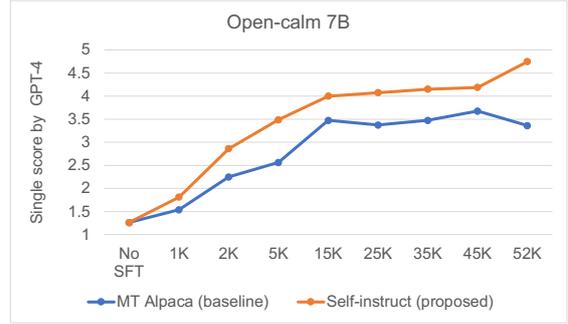

Figure 6: The Open-calm 7B learning Curve

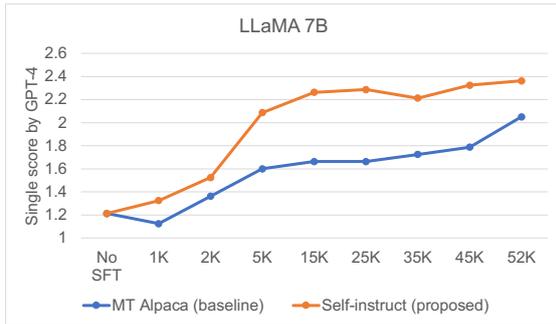

Figure 5: The LLaMA 7B learning Curve

Furthermore, it is worth noting that as the amount of data increases, the scores of the model answer will increase smoothly and finally converge to a constant value. In contrast, an increase in the amount of data from $1K$ to $5K$ resulted in a significant increase in the model answer score. And the model answer score is optimal when the amount of instruction data increases to $5K$. We argue that the instruction data used to SFT the language model, while ensuring high quality, requires only a small amount of data to improve the model's capabilities and performance dramatically.

Meanwhile, comparing two sets of instruction data, our dataset substantially improves the models' performance compared to the MT Alpaca with a large margin at each checkpoint, confirming the high-quality instruction data generated by our proposal. Subsequently, just $5K$ self-instruct data leads to a competitive performance with the whole $5K$ MT Alpaca data, indicating that the key to improving the effectiveness of the model is through a small amount of high-quality data as opposed to boosting the quantity of data.

With that conclusion, we only need to manually translate the seed task and then generate a small amount of instruction data through self-instruction to constitute high-quality instruction dataset for a language.

### 5.2. Fine-tuned Model Performance Compared to GPT3.5

#### 5.2.1. Single Score Mode

| Base model | Instruction | Score |
|---|---|---|
| Davinci-003 | - | 5.86 |
| LLaMA 7B | MT Alpaca | 2.05 |
| LLaMA 7B | Self-instruct | 2.36 |
| LLaMA2 7B | MT Alpaca | 4.45 |
| LLaMA2 7B | Self-instruct | 5.71 |
| Open-calm 7B | MT Alpaca | 3.36 |
| Open-calm 7B | Self-instruct | 4.75 |
| LLaMA2 13B | MT Alpaca | 5.30 |
| LLaMA2 13B | Self-instruct | **6.06** |

Table 1: Single scores of models with $52K$ data. The underline denotes the higher score between two sets of instruction data.

In this section, we employed the single mode to evaluate all the models fine-tuned with $52K$ instruction data with a single score of quality judged by GPT-4, including GPT-3.5 (Davinci-003). As shown in Table 1, for all four base models, fine-tuning the self-instruct data significantly outperforms MT Alpaca, which validates our claim that, due to GPT-4 and the omission of the translation step, the quality of our self-instruct data significantly surpasses MT Alpaca. An interesting observation is that stronger models benefit more from our proposal. For instance, LLaMA2 obtains $1.26$ points increase and Open-calm obtains $1.39$ points increase, both of which substantially surpass the $0.31$ points increase of LLaMA.

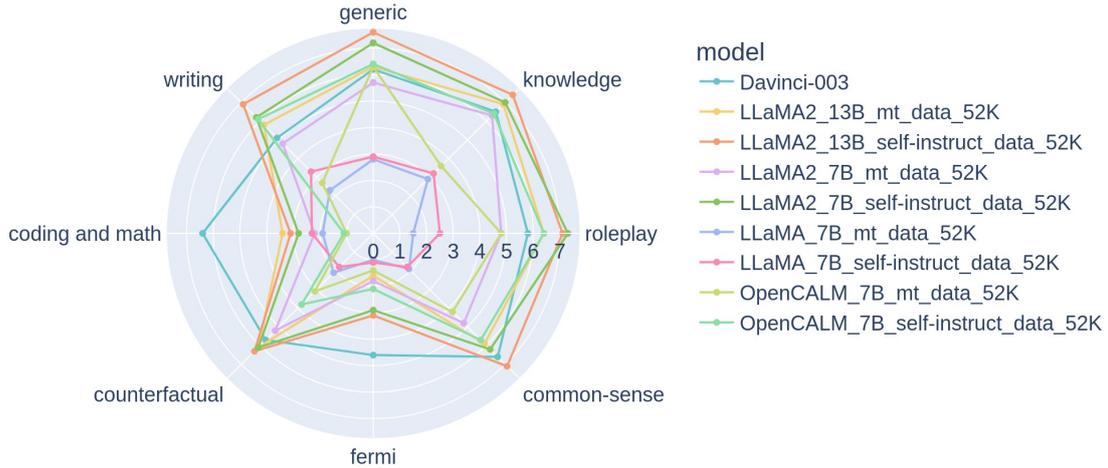

Figure 7: Single score performance of $52K$ fine-tuned models and GPT-3.5 (davinci-003) of each question category

| Base model | Instruction | Win-rate |
|---|---|---|
| LLaMA 7B | MT Alpaca | 5.99 |
| LLaMA 7B | Self-instruct | <u>13.12</u> |
| LLaMA2 7B | MT Alpaca | 33.12 |
| LLaMA2 7B | Self-instruct | <u>46.25</u> |
| Open-calm 7B | MT Alpaca | 15.62 |
| Open-calm 7B | Self-instruct | <u>34.37</u> |
| LLaMA2 13B | MT Alpaca | 32.50 |
| LLaMA2 13B | Self-instruct | **54.37** |

Table 2: Win-rate compared with GPT-3.5 with $52K$ instruction data. The underline denotes the higher win-rate between two sets of instruction.

After being fine-tuned with the self-instruct data, LLaMA2 7B performs very close to GPT-3.5 and 13B finally outperforms GPT-3.5 by $0.2$ points. This demonstrates the great potential of our proposal in the language of Japanese, enabling smaller LLMs to approach the similar generation quality to GPT-3.5.

#### 5.2.2. Pairwise Win-rate Mode

In this section, we employed the pairwise mode to compare each fine-tuned LLM to the GPT-3.5 (Davinci-003). GPT-4 judges the answer pairs and assigns 'win/lose/tie' against GPT-3.5. The results are finally converted into a win-rate score for each model according to the following equation.

$$\text{win-rate} = \frac{\text{win} + \frac{\text{tie}}{2}}{\text{win} + \text{loss} + \text{tie}}$$

In Table 2, a similar observation is found that the LLMs fine-tuned with the self-instruction data achieve higher win-rate scores against GPT-3.5. LLaMA2 7B can achieve $46.25\%$ win-rate against GPT3.5 and the 13B model significantly beats it with a $54.37\%$ win-rate.

In addition, based on the data in Table 2 and Table 1, it is apparent that there is significant consistency between the single score mode and pairwise win rate mode. The models that evaluated with superior scores in single score mode also yield higher win-rate against GPT-3.5.

#### 5.2.3. Scores on Each Question Category

Figure 7 shows that LLMs, when SFT with self-instruct data perform competitively or even better in most categories. Especially, LLaMA2 SFT with $52K$ self-instruct data achieves a higher score than GPT-3.5 davinci-003 in all categories except 'coding and math' and 'fermi'.

A tentative hypothesis could be that a sufficient size is necessary for answering both categories, considering the emergent ability of LLMs (Wei et al., 2022b), while the size of 13B limits our models.

Another possible reason could be the deficiencies of generated instruction data involving these two categories, leading to the poor performance compared with GPT-3.5.

| Quality Category | Sample Instruction | Sample Input | Sample Output | Proofread Sample |
|---|---|---|---|---|
| High quality | 入力された文章を、5歳の子供が分かるように単純化してください。(Simplify the input text so that a 5-year-old child can understand it.) | 人類は標準的な銀河間航行テクノロジーをまだ発見していません。(Humanity has not yet discovered standard intergalactic navigation technology.) | まだ人間は宇宙の遠くの星に行く方法を見つけていません。(As yet, humans have not found a way to reach the distant stars of the universe.) | – |
| Non-fluent text | 「与えられた指示の順序を並び替えて、<span style="color:red">有効なコマンドの順序</span>を作成してください。」("Rearrange the order of the given instructions to create a <span style="color:red">valid sequence of commands</span>.") | 新しいユーザーアカウントを追加し、ログインしてパスワードを割り当てる (Add a new user account, log in and assign a password.) | 「ログインし、新しいユーザーアカウントを追加し、パスワードを割り当てる。」("Log in, add a new user account, and assign a password.") | 適切な指示 (appropriate instructions) |
| Format violation | フレーズを並列構造を使って書き直しなさい。(Rewrite the phrases using parallel structures.) | パイロットの仕事は、飛行機を操作して飛ばすことです。(A pilot's job is to operate the airplane for flying.) | パイロットの仕事は、<span style="color:red">飛行機を操作して飛行すること</span>です (A pilot's job is <span style="color:red">to operate the airplane for flying</span>.) | 飛行機を操作することと飛ばすことです。(to operate and fly airplanes.) |

Table 3: Instruction data samples in three quality categories. Red denotes the problematical content.

| Dataset | HQ | NT | FV |
|---|---|---|---|
| MT Alpaca | 42 | 28 | 30 |
| Self-instruct (ours) | 67 | 27 | 6 |

Table 4: Manual check of two types of instruction data. HQ, NT, and FV denote high quality, non-fluent text, and format violation, respectively.

## 6. Additional analysis

### 6.1. Manual Instruction Data Quality Check

We conducted a quality assessment of the instruction data by randomly selecting $100$ samples each from the self-instruction and machine-translated data. We then manually sorted the samples into three quality categories:

- **Format violation**: An instruction text is non-instructive or a question/answer text does not follow or is not coherent to an instruction.

- **Non-fluent text**: Except for the above, an instruction, question, or answer text is syntactically or semantically incorrect or unnatural.

- **High quality**: An instruction, question, and answer texts adhere to the format of instruction data and are both fluent and natural.

Note that we did not focus on the exact correctness of the answers. This is because the primary objective of instruction data is to train the model to follow instructions, not to provide it with exact factual information.

In Table 3, we give an example for each quality category. In the high-quality example, the instruction data does not contain obvious problematic content. In the non-fluent example, the red part '有効なコマンドの順序' means 'a valid command sequence.' However, in Japanese, it is preferred to use '適切な指示 (appropriate instructions)' rather than 'valid command sequence' in this context. In the format violation example, the instruction requests to 'rewrite the phrase using parallel structures.' However, the output section does not follow the instruction at all.

| category | win | loss | tie |
|---|---|---|---|
| generic | 3 | 2 | 5 |
| knowledge | 4 | 1 | 5 |
| roleplay | 3 | 3 | 4 |
| common-sense | 4 | 1 | 5 |
| fermi | 2 | 1 | 7 |
| counterfactual | 5 | 0 | 5 |
| coding | 3 | 0 | 4 |
| math | 0 | 0 | 3 |
| writing | 7 | 0 | 3 |
| **Total** | **31** | **8** | **41** |

Table 5: The human evaluation results of LLaMA2 13B with self-instruct data vs LLaMA2 13B with MT Alpaca data. Both models use $52K$ instructions.

Table 4 clearly demonstrates that the quality of self-instruct data is significantly higher than that of MT Alpaca data. However, $6$ samples from the self-instruction data still exhibit format-violations.

### 6.2. Manual Evaluation of The Model Answers

Human evaluation was conducted between LLaMA2 13B fine-tuned with the self-instruct data and with the MT Alpaca data. The evaluation followed the same method as the pairwise mode evaluation of comparing two models and judging a win, loss, or tie result. The human evaluation results are shown in Table 5.

Across all categories, the self-instruction data outperforms the MT Alpaca data. The human evaluation results are consistent with the GPT-4 evaluation presented in Section 5.1 and 5.2. It underscores the significance of high-quality self-instruct instruction data.

When analyzed by category, the most substantial improvement was observed in the counterfactual category. In answers to counterfactual questions, the self-instruction data can provide greater depth and detail. On the other hand, roleplay and math categories displayed no noticeable enhancements. In the case of math, both provided incorrect arithmetic results for all questions. It suggested that the quality of instruction has less influence on arithmetic performance.

### 6.3. Ablation Study

It is worth noting that in our study, we use GPT-4 to generate Japanese instruction data, while in the baseline work, they translate the English data generated by GPT-3.5. This difference leaves a concern that we are not sure which factor leads to the performance improvements most, whether it is the performance gap between GPT-4 and GPT 3.5 or the difference made by the proposed approach.

In order to address this concern, we designed the following ablation study. We randomly sampled $5K$ samples from the Alpaca GPT-4 instruction data (Peng et al., 2023). This English instruction-following data is generated by GPT-4 using Alpaca prompt for fine-tuning LLMs. Then we translated it into Japanese by using DeepL. Also, we randomly sampled 5K samples from our Japanese Self-instruct dataset. We then perform SFT on LLaMA2 7B using these 2 datasets.

- Self-instruct ($5K$): $5K$ instruction data generated by our approach.

- MT Alpaca ($5K$): $5K$ GPT-4 Alpaca instruction data machine translated by DeepL.

From Table 6, we can see that even though the original English dataset is also generated by GPT-4, the translated baseline performs significantly worse than our proposal. The result demonstrates that the machine translation process does lead to the deterioration in data quality and our proposal can effectively can effectively avoid such deterioration.

| Proposal | GPT-4 baseline | Win-rate |
|---|---|---|
| Self-instruct ($5K$) | MT Alpaca ($5K$) | 55.6 |

Table 6: We separately STF LLaMA2 7B with 2 instruction datasets. This is the Win-rate for the proposal directly compared to translated GPT-4 Alpaca.

## 7. Conclusion

This paper introduces an efficient paradigm for developing resources in non-English languages like Japanese with minimal human effort. By translating a small set of English instructions into Japanese and subsequently post-editing them for native-level quality, we enable GPT-4 to generate Japanese instruction data. Besides, we construct an evaluation benchmark with 80 questions across eight categories, using GPT-4 to assess large language models without human references automatically. Experiment results demonstrate that models fine-tuned on our GPT-4 self-instruct data consistently outperform existing approaches, and human evaluations validate the consistency between GPT-4's assessments and human preferences, underscoring the promise of our methodology for advancing large language models in non-English contexts. Additionally, during the analysis, we also found that the quality of instruction data shows more significance than the quantity of the data, which may further guide the study in the instruction tuning field.

## 8. Acknowledgement

This work was supported by Cross-ministerial Strategic Innovation Promotion Program (SIP) on "Integrated Health Care System" Grant No. JPJ012425, JSPS KAKENHI Grant No. JP23K16946, JSPS KAKENHI Grant No. JP23H03454 and the AI fellowship from Kyoto University.

We specially thank Hirokazu Kiyomaru for his excellent efforts to improve and maintain our Japanese Vicuna QA Benchmark repository.

## 9. Bibliographical References

## 10. Language Resource References